\documentclass{article}

    \PassOptionsToPackage{numbers, compress}{natbib}

\usepackage[preprint]{neurips_2025}

\usepackage[utf8]{inputenc} 
\usepackage[T1]{fontenc}    
\usepackage{hyperref}       
\usepackage{url}            
\usepackage{booktabs}       
\usepackage{nicefrac}       
\usepackage{microtype}      
\usepackage{amsmath}
\usepackage{graphicx}
\usepackage{subfig}
\usepackage{float}
\usepackage{tabularx} 
\usepackage{mlmath}
\usepackage{longtable} 
\usepackage[table]{xcolor} 
\usepackage{multirow} 

\usepackage{natbib}

\definecolor{pcolor}{HTML}{00C562}
\definecolor{ppcolor}{HTML}{00774F}
\definecolor{mcolor}{HTML}{DD4F4F}
\newcommand{\pcolor}[1]{{\color{pcolor}{#1}}}
\newcommand{\ppcolor}[1]{\textbf{\color{ppcolor}{#1}}}
\newcommand{\mcolor}[1]{{\color{mcolor}{#1}}}

\usepackage{threeparttable}
\usepackage{threeparttablex}
\usepackage{amsmath,amsfonts,amssymb,mathtools,amsthm}

\newtheorem*{theorem*}{Theorem}

\newtheorem*{lemma*}{Lemma}

\newtheorem*{property*}{Property}

\newtheorem*{assumption*}{Assumption}
\newtheorem*{prop*}{Proposition}

\newtheorem*{setting*}{Setting}

\newcommand{\R}{\mathbb{R}}
\newcommand{\x}{\mathbf{x}}
\newcommand{\w}{\mathbf{w}}

\title{Scalable Complexity Control Facilitates Reasoning Ability of LLMs}

\author{%
    Liangkai Hang\textsuperscript{\rm 1, $\dag$},
    Junjie Yao\textsuperscript{\rm 1, $\dag$},
    Zhiwei Bai\textsuperscript{\rm 1},
    Tianyi Chen\textsuperscript{\rm 1},
    Yang Chen\textsuperscript{\rm 1},
    Rongjie Diao\textsuperscript{\rm 1},\\
    {\bf Hezhou Li\textsuperscript{\rm 1}},
    {\bf Pengxiao Lin\textsuperscript{\rm 1}},
    {\bf Zhiwei Wang\textsuperscript{\rm 1}}, 
    {\bf Cheng Xu\textsuperscript{\rm 1}}, 
    {\bf Zhongwang Zhang\textsuperscript{\rm 1,*}}, 
    {\bf Zhangchen Zhou\textsuperscript{\rm 1}},\\ 
    {\bf Zhiyu Li}\textsuperscript{\rm 4,5,*},
    {\bf Zehao Lin}\textsuperscript{\rm 4,5},
    {\bf Kai Chen}\textsuperscript{\rm 5},
    {\bf Feiyu Xiong}\textsuperscript{\rm 4,5, *},\\
    {\bf Yaoyu Zhang}\textsuperscript{\rm 1,2, *},
    {\bf Weinan E}\textsuperscript{\rm 3,4},
    {\bf Hongkang Yang}\textsuperscript{5,*},
    {\bf Zhi-Qin John Xu\textsuperscript{\rm 1,2,4}\thanks{Corresponding author: xuzhiqin@sjtu.edu.cn,hongkang@alumni.princeton.edu,zhyy.sjtu@sjtu.edu.cn, xiongfy@iaar.ac.cn, lizy@iaar.ac.cn, 0123zzw666@sjtu.edu.cn}} \\
    \textsuperscript{\rm
 1} Institute of Natural Sciences, School of Mathematical Sciences, Shanghai Jiao Tong University\\
    \textsuperscript{\rm 2} MOE-LSC, School of Artificial Intelligence, Shanghai Jiao Tong University\\
    \textsuperscript{\rm 3} Center for Machine Learning Research, School of Mathematical Sciences, Peking University\\
    \textsuperscript{\rm 4} Center for LLM, Institute for Advanced Algorithms Research, Shanghai\\
    \textsuperscript{\rm 5} MemTensor (Shanghai) Technology Co., Ltd.\\
    \textsuperscript{\rm $\dag$} Equal contribution, list in alphabetical order.
}

\begin{document}

\maketitle

\begin{abstract}
The reasoning ability of large language models (LLMs) has been rapidly advancing in recent years, attracting interest in more fundamental approaches that can reliably enhance their generalizability.
This work demonstrates that model complexity control, conveniently implementable by adjusting the initialization rate and weight decay coefficient, improves the scaling law of LLMs consistently over varying model sizes and data sizes.
This gain is further illustrated by comparing the benchmark performance of 2.4B models pretrained on 1T tokens with different complexity hyperparameters.
Instead of fixing the initialization std, we found that a constant initialization rate (the exponent of std) enables the scaling law to descend faster in both model and data sizes.
These results indicate that complexity control is a promising direction for the continual advancement of LLMs.
\end{abstract}

\section{Introduction}





\begin{figure}[htpb]
    \vspace{-8pt}
    \centering
    \includegraphics[width=1\linewidth]{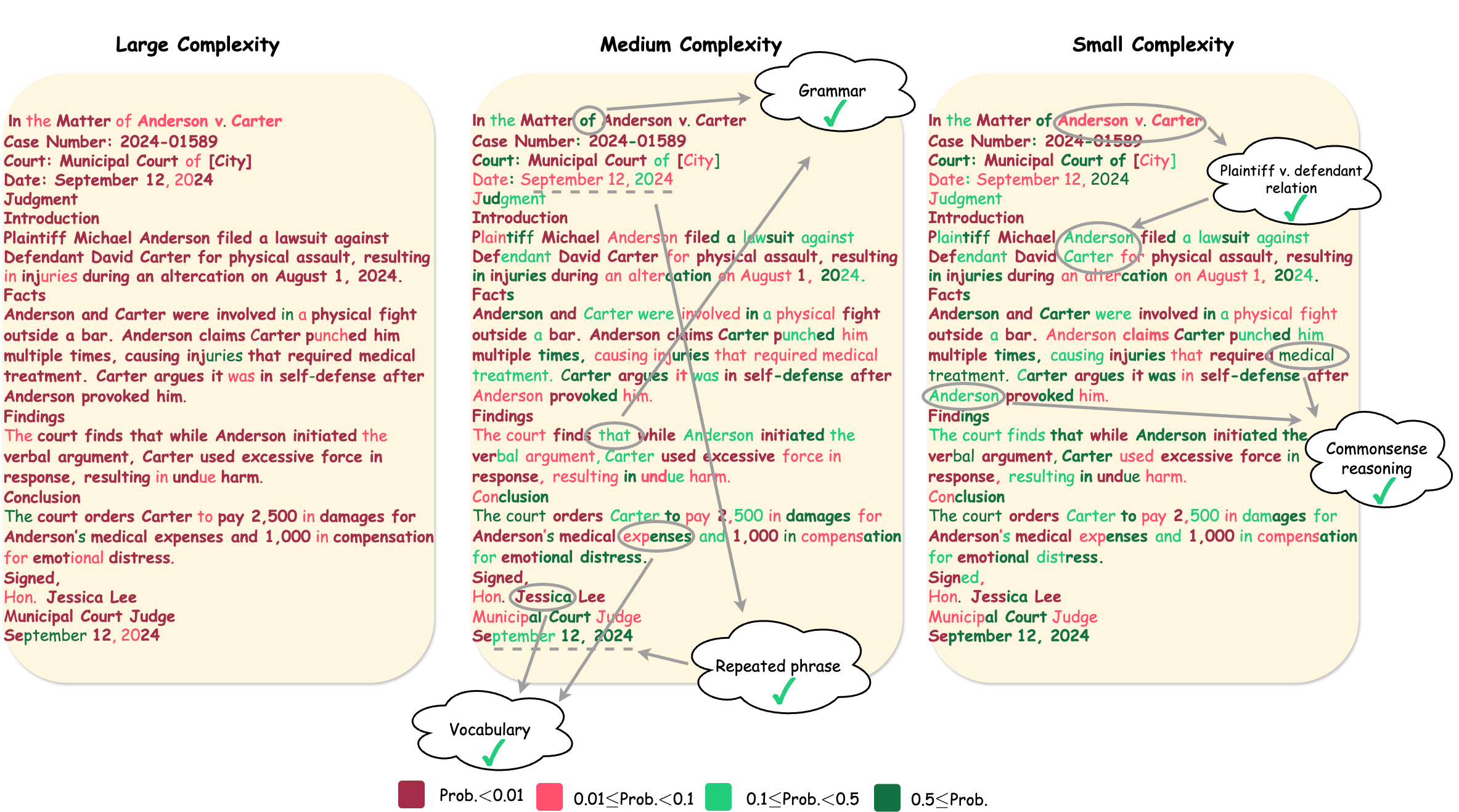}
\caption{Next-token prediction accuracy with varying model complexity.
The colors indicate the probability of each token predicted by the models
Model complexity decreases from left to right, as the initialization rate $\gamma=0.1,0.5,1$ (see Section \ref{sec:cc} for the definition of $\gamma$).
These models have the same shape (180M parameters) and are trained the same dataset (40B tokens).}
\label{fig:token_loss}
\end{figure}\vspace{-5pt}


In recent years, large language models (LLMs) have achieved unprecedented progress, demonstrating impressive performance on a wide range of tasks~\cite{achiam2023gpt,liu2024deepseek,touvron2023llama,wei2022emergent,yang2024qwen2,zeng2021pangu}.
The key to this success is the improved generalizability of these models, in particular, their reasoning ability.
Various approaches have been explored to enhance reasoning, such as post-training with reinforcement learning~\cite{kumar2024training,guo2025deepseek},
high-quality math and code data with reasoning traces \cite{guo2025deepseek},
chain-of-thought and related prompting strategies~\cite{wei2022chain},
and the separation of reasoning and memory in pretraining \cite{yang2024memory3}.


This work identifies model complexity as the key factor in the development of the reasoning ability of LLMs.
Complexity control can be implemented through various means, such as adjusting the rate of parameter initialization ~\cite{luo2021phase,zhou2022empirical,zhou2022towards,zhang2025complexity,yao2025analysis} or applying stronger penalties on parameter norms ~\cite{zhang2025complexity}.
Similar to the gene that determines the characteristics of an organism, these designs directly regulate the reasoning ability of LLMs.

An illustration is provided in Figure~\ref{fig:token_loss}.
On the left, a pretrained model with large complexity (large initialization scale) fails to make meaningful next-token predictions on unseen test data.
In the middle, with moderate complexity (the commonly used initialization scale), the model demonstrates basic knowledge of grammar (e.g. ``finds \underline{that}'') and vocabulary (e.g. ``Jess\underline{ica}''), as well as the induction head mechanism \cite{olsson2022incontext} (e.g. the repeated phrase ``September \underline{12, 2024}'').
On the right, the low-complexity model  (i.e. with small initialization scale) further captures sophisticated semantic reasoning,
successfully predicting ``Plaintiff Michael \underline{Anderson}'' given the context ``Anderson v. Carter'' despite that the full name ``Michael Anderson'' has not appeared before,
and also predicting ``required \underline{medical treatment}'' given that Anderson was punched by Carter.

Intuitively, smaller complexity forces the model to compress data into a smaller set of production rules, revealing the deep dependency among the tokens and preventing plain memorization.
Previous studies \cite{luo2021phase,zhou2022empirical} have shown that with a smaller initialization scale, neurons within each layer tend to evolve within a few groups,
a phenomenon known as condensation, which limits the effective number of neurons. Readers are referred to an overview of condensation \cite{xu2025overview}.
Likewise, models trained with stronger penalty on parameter norm may converge to solutions with smaller norms, resulting in lower-complexity outputs.
Thus, it is promising that these techniques can enhance the reasoning ability of LLMs.


As a verification of this intuition, our experiments exhibit improved scaling laws in data size and model size, as well as higher scores over almost all benchmarks (+4.6\% for 0.9B model with 600B data, and +3.4\% for 2.4B model with 1T data, averaged over 15 tasks).
Compared with the standard deviation (std) of parameter initialization, the initialization rate (the exponent of std as a function of network width) turns out to be the right invariant for the scaling laws.
This is in accordance with previous works \cite{luo2021phase,zhou2022empirical} on the phase diagram of neural network training.
We provide some heuristic calculations to explain how complexity control facilitates the learning of multi-step reasoning.

\section{Related works}

\paragraph{LLMs reasoning ability}
Even advanced LLMs such as GPT-4 often struggle with implicit reasoning over parametric knowledge~\citep{talmor2020olmpics, kassner2020pretrained, rae2021scaling, press-etal-2023-measuring, allenzhu2023physics, yang2024large}, revealing their limited ability to internalize structured facts and rules. Verbalized reasoning strategies such as chain-of-thought can substantially boost performance, particularly for large models~\citep{wei2022chain, zelikman2022star, sun2023recitationaugmented, liu-etal-2023-crystal, zelikman2024quietstar, feng2023towards, li2024chain}.
However, understanding the underlying capacity for implicit reasoning remains a critical challenge, often studied by controlled experiments~\citep{prystawski2023why, dziri2023faith, wang2024understanding}.
This work takes the more intrinsic perspective of complexity control.

\paragraph{Effect of initialization}
Parameter initialization is known to be influential to the training and performance of classical neural networks
\cite{arora2019exact, chizat_global_2018, zhang_type_2019, e2020comparative, jacot_neural_2018, mei_mean_2018, rotskoff_parameters_2018, sirignano_mean_2020, williams_gradient_2019}.
For instance, distinct phases (the linear and condensed regimes) can be induced in wide ReLU networks by varying initialization rates
\cite{luo2021phase,zhou2022empirical},
and training in the condensed regime tend to fit data with lower-complexity functions~\cite{zhang2022linear, zhang2023loss, zhang2023stochastic, zhang2024implicit}.
The particular case of Transformer language models has also been studied
\cite{huang2020improving, liu2020understanding, trockman2023mimetic, wang2024deepnet, zhang2019improving, zhu2021gradinit, zhang2024initialization, zhang2025complexity},
with particular interest on the impact of initialization on training stability and efficiency.
Experiments on toy datasets~\cite{zhang2024initialization, zhang2025complexity} show that small initialization scales assist Transformers to identify the elementary functions when fitting synthetic compositional data.
However, the influence of initialization on the reasoning ability of Transformers trained on natural language data remains to be explored.

\paragraph{Weight decay}
\cite{krogh1991simple} introduced weight decay as a method to improve the generalization of neural networks.
Many subsequent works have explored its role in controlling model complexity and enhancing generalization~\cite{ bartlett2002rademacher,bartlett2017spectrally,arora2018stronger,neyshabur2015norm,golowich2018size,wei2019data}.
More recently, weight decay has been shown to be particularly important for achieving better generalization~\cite{power2022grokking, varma2023explaining}.

In this work, we systematically investigate how controlling pre-training complexity, via initialization strategies and weight decay, affects the performance of LLMs. Unlike prior studies that focus on training stability or synthetic tasks, we evaluate the impact across a wide range of downstream benchmarks and analyze the underlying mechanisms, aiming to offer practical guidance for large-scale model pre-training.

\section{Complexity control}\label{sec:cc}
In this section, we introduce the approach to modulate the model complexity.
 \paragraph{Initialization rate $\gamma$} Given any trainable parameter matrix $\vW\in\sR^{d_{in}\times d_{out}}$. We initialize its elements according to the following normal distribution:
\begin{equation*}
    \vW_{i,j} \sim \mathcal{N}\left(0,\left(d_{in}^{-\gamma}\right)^2\right),
\end{equation*}
where $\gamma$ is the initialization rate. Specifically, the initialization scale decreases as $\gamma$ increases.
Note that $\gamma=0.5$ is commonly used in many default initialization methods, such as LeCun initialization~\cite{LeCun1998} and He initialization~\cite{he2015delving}. As the network width towards infinity~\cite{luo2021phase,zhou2022empirical}, the training of the network with $\gamma>0.5$ exhibits significant non-linear characteristics, i.e., condensation. Therefore, initialization scales with $\gamma>0.5$ are generally considered small. Tuning $\gamma$ is a scalable approach for complexity control. 

\paragraph{Weight decay coefficient $\lambda$}Given any trainable parameter $\theta_t$ where $t$ denotes the current training step. Define $\hat{\theta}_t$ as the parameter after optimizing by gradient and moment, the weight decay is implemented by
\begin{equation}
    \theta_{t+1} \longleftarrow \hat{\theta}_t - \lambda C \theta_t,
\end{equation}
where $\lambda$ is the weight decay coefficient.

\section{Results}
To evaluate the impact of complexity control, we train LLMs based on the Llama- architecture~\cite{touvron2023llama} under different levels of model complexity. We examine scaling laws with respect to both model size and training data size, and compare different models across a range of benchmarks. Detailed setup of training and evaluation is provided in Appendix~\ref{app:setup}.


\subsection{Scaling law}
\begin{figure}[htpb]
    \centering
    \includegraphics[width=1\linewidth]{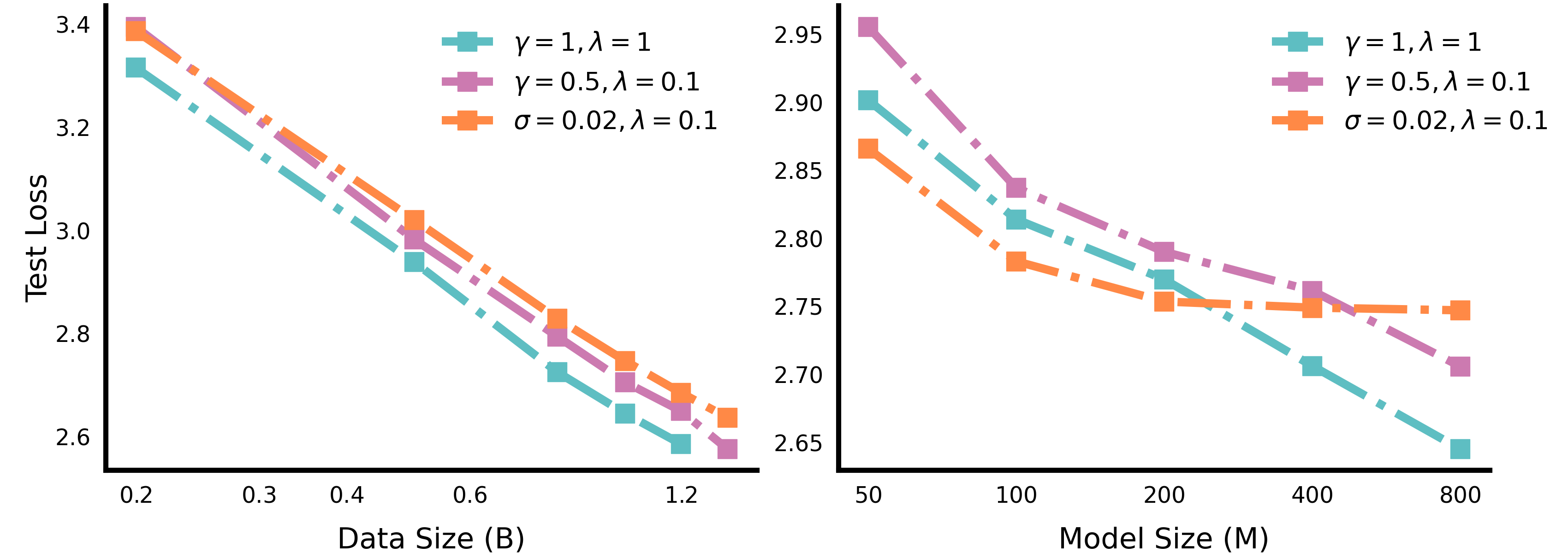}
    \caption{Test loss across varying data and model scales under different complexity configurations. Left: Test loss progression for 0.8B-parameter models trained with data scales ranging from 0.2B to 1.4B tokens. Right: Test loss versus model parameter counts (50M-0.8B) with fixed 1B training tokens. Line colors correspond to different complexity configurations. }
    \label{fig:scalinglaw}
\end{figure}

We first establish three model complexity configurations: (1) the small-complexity setup with $\gamma=1,\lambda=1$, (2) the large-complexity setup with $\gamma=0.5,\lambda=0.1$, and (3) the commonly used default configuration (e.g., GPT2, HuggingFace) with $\sigma=0.02,\lambda=0.1$. Under each configuration, we train 0.8B-parameter models with varying training data scales ranging from 0.2 billion to 1.4 billion tokens. Figure~\ref{fig:scalinglaw} (left) demonstrates the relationship between test loss and data scale across different complexity configurations. The curve of small-complexity (green) exhibits a distinct leftward shift relative to the large-complexity (purple), suggesting that complexity controlling can effectively improve the sample efficiency. Additionally, while the test loss of the large-complexity model demonstrates comparable to the standard configuration at smaller data scales, their performance diverges as data size increases. Specifically, the former configuration achieves progressively lower test loss than the latter, indicating superior scalability of the $\gamma$-initialization approach with steeper scaling slopes. Besides, we train models with distinct model sizes by 1 billion tokens. Figure~\ref{fig:scalinglaw} (right) reveals similar patterns in the relationship between test loss and model size. Although $\sigma$-initialization achieves lower test loss with smaller models, its performance plateaus as the model size increases, resulting in significantly higher loss compared to $\gamma$-initialized models at larger scales. These results illustrate the scalability potential of the $\gamma$-initialization method, maintaining performance advantages across expanding parameters and data scales. 
\vspace{-15pt}
\subsection{Evaluation}
\begin{table}[htbp]
\centering
\begin{threeparttable}
\caption{Evaluation of models with different model complexities.}
\label{tab:model_benchmark}
\begin{tabular}{@{}lccccc@{}}
\toprule
\textbf{Models\tnote{1}} & \textbf{0.9B Large} & \textbf{0.9B Small} & \textbf{2.4B Large} & \textbf{2.4B Small}\\
$\left(\gamma,\lambda\right)$ & (0.5,0.1) &(1,1) &(0.5,0.1) & (0.58,1)\\
\midrule
MMLU          & 49.9  & 52.5 (\pcolor{+2.6})  & 60.4  & 64.4 (\ppcolor{+4}) \\
MMLU-Pro      & 17.6  & 21.5 (\pcolor{+3.9})  & 30.6  & 30.1 (\mcolor{-0.5}) \\
BBH           & 33.3  & 34.9 (\pcolor{+1.6})  & 42.4  & 43.7 (\pcolor{+1.3}) \\
ARC-C         & 46.1  & 49.8 (\pcolor{+3.7})  & 58.6     & 60.7 (\pcolor{+2.1}) \\
TruthfulQA    & 53.4  & 56.8 (\pcolor{+3.4})  & 58.7     & 61.7 (\pcolor{+3}) \\
WinoGrande   & 68.0  & 72.2 (\ppcolor{+4.2}) & 73.1 & 76.9 (\pcolor{+3.8}) \\
HellaSwag     & 63.0  & 67.3 (\ppcolor{+4.3}) & 71.2     & 75.4 (\ppcolor{+4.2}) \\
AGIEval-EN    & 25.3 & 29.1 (\pcolor{+3.8}) & 33.2 & 35.6 (\pcolor{+2.4})\\

OpenBookQA    &38.8  & 39.6 (\pcolor{+0.8}) & 41.8  & 43.4 (\pcolor{+1.6}) \\

CommonsenseQA  &57.7  & 67.2 (\ppcolor{+9.5}) &70.4  & 76.9 (\ppcolor{+6.5})\\

GPQA          & 25.9  & 26.8 (\pcolor{+0.9})  & 29.0     & 31.7 (\pcolor{+2.7}) \\
MATH          & 5.9  & 11.5 (\ppcolor{+5.6})  & 34.3  & 35.3 (\pcolor{+1}) \\
GSM8K         & 21.8  & 40.2 (\ppcolor{+19.4}) & 52.8  & 63.8 (\ppcolor{+11.0}) \\

MBPP          & 6.6  & 10.8 (\ppcolor{+4.2})  & 20.6  & 22.4 (\pcolor{+1.8})  \\
IFEval        & 31.3  & 32.0 (\pcolor{+0.7}) & 34.7  & 40.9 (\ppcolor{+6.2})\\
\bottomrule

\end{tabular}
\begin{tablenotes}
    \footnotesize
        \item[1] ``Large'' and ``Small'' mean large complexity and small complexity, respectively. 
\end{tablenotes}
\end{threeparttable}
\end{table}
To assess the impact of complexity control, we adopt reasonable complexity configurations and train LLMs with the following setup: (1) 0.9B-parameter models on 600B tokens from SlimPajama~\cite{cerebras2023slimpajama} and (2) 2.4B-parameter models trained with 1T high-quality corpus. We evaluate the performance of our models across a comprehensive set of benchmark tasks. As shown in Table~\ref{tab:model_benchmark}, the performance of small-complexity models improves significantly. Specifically, complexity controlling yields substantial gains in reasoning capabilities. On math-related tasks, two small-complexity models achieve improvements of 19.4 and 11.0 points on the GSM8K benchmark, respectively, as well as 5.6 and 1 points on the MATH dataset. Moreover, these small-complexity models demonstrate notable performance enhancements across other reasoning tasks, including Winogrande($+4.2,+3.8$), HellaSwag($+4.3,+4.2$), and CommonsenseQA($+9.5,+6.5$). These results demonstrate that principled control of model complexity can significantly enhance the overall capabilities of LLMs, particularly in reasoning tasks. Note that small initialization may yield slower training. For 2.4B model, the performance of $\gamma=0.58$ is slightly better than that of $\gamma=1$ in Table \ref{tab:0.58-1} in Appendix.

Complexity control also significantly improves the performance of base models, with results summarized in Table~\ref{tab:base_benchmark} in Appendix~\ref{app:base}.  
As quantified in Figure~\ref{fig:base_sft}, the small-complexity model attains greater score increments across most tasks at the SFT stage. 
\begin{figure}[htpb]
    \centering
    \includegraphics[width=0.9\linewidth]{ 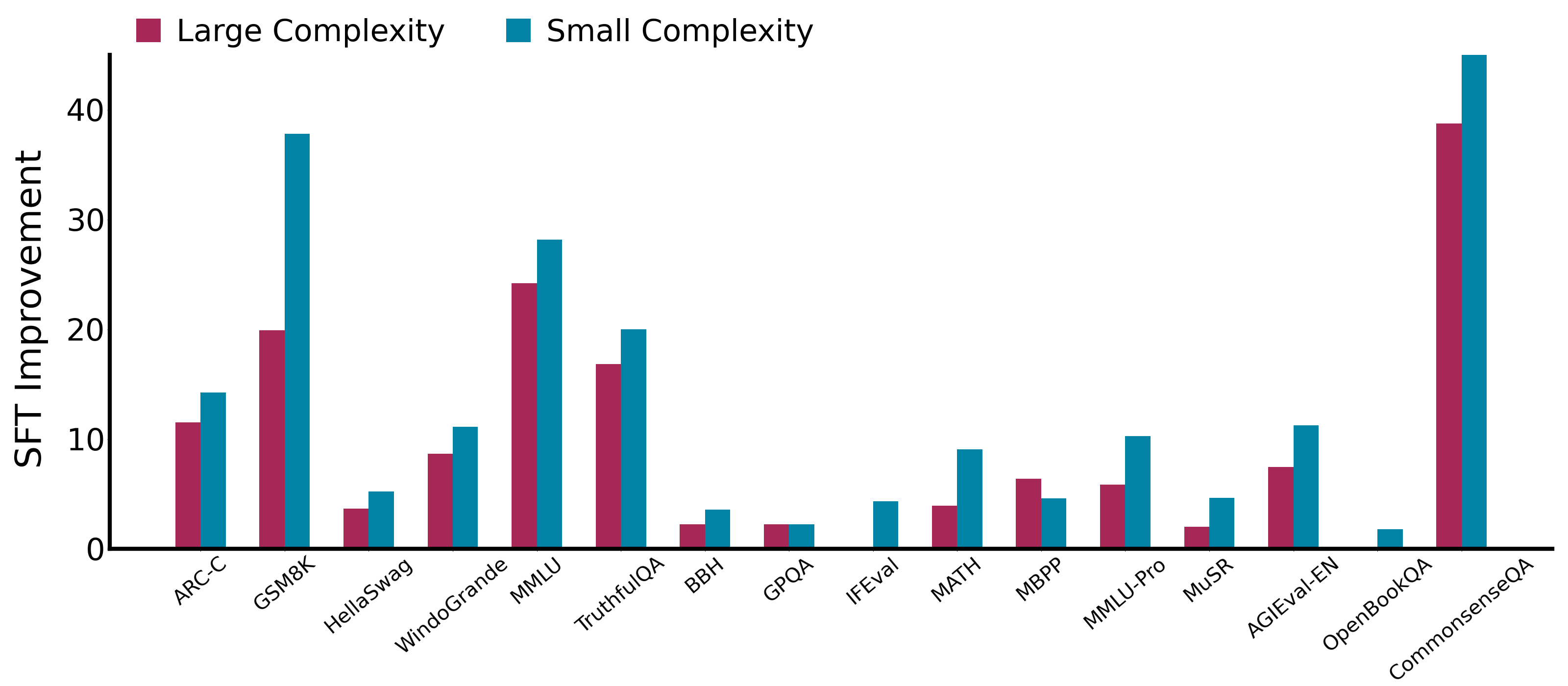}
    \caption{Performance improvement via SFT across complexity configurations(0.9B model). It is quantified as the performance gap between the SFT model and the corresponding base model.}
    \label{fig:base_sft}
\end{figure}
\vspace{-8pt}

\section{Analysis}\label{sec:analysis-187}
To investigate the effect of initialization scale and weight decay in complexity control, we trained 180M-parameter models with different complexity configurations. We conduct analysis from evaluation and parameter analysis, which yield mechanistic insights into complexity control.

\subsection{Influence of initialization scale and weight decay}
\begin{figure}[htpb]
    \centering
    \includegraphics[width=1\linewidth]{ 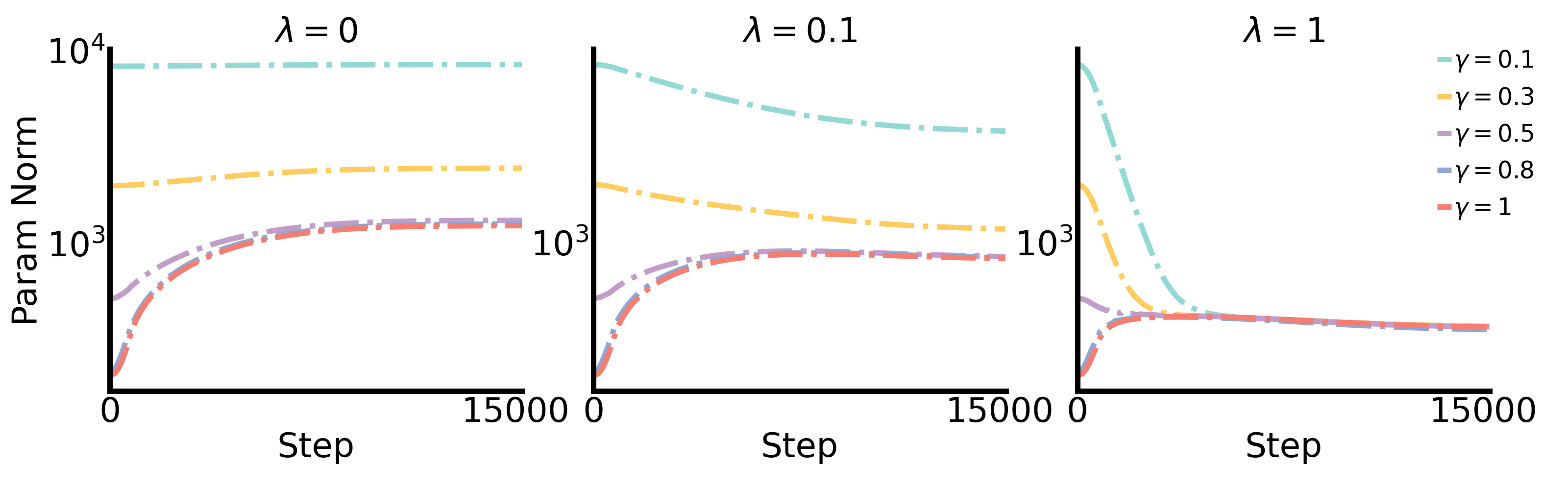}
    \caption{Parameter norm evolution across complexity configurations. Left to right: $\lambda = 0, 0.1, 1$; Line colors correspond to $\gamma = 0.1,0.3, 0.5,0.8,1$.}
    \label{fig:paranorm}
\end{figure}
Examining the interplay between initialization scale $\gamma$ and weight decay $\lambda$ on model complexity is critical. Figure~\ref{fig:paranorm} visualizes the temporal evolution of parameter norms under varying $\gamma-\lambda$, demonstrating that large $\gamma$ coupled with large $\lambda$ systematically induces lower model complexity. Specifically, with small $\lambda$ ($\lambda=0,0.1$), configurations with larger $\gamma$ converge to smaller complexities. However, with large $\lambda$ ($\lambda=1$), models with different $\gamma$ converge to comparable small norm.

\subsection{Alignment between complexity and capability}\label{sec:187m}
\begin{figure}[htpb]
    \centering
    \includegraphics[width=1\linewidth]{ 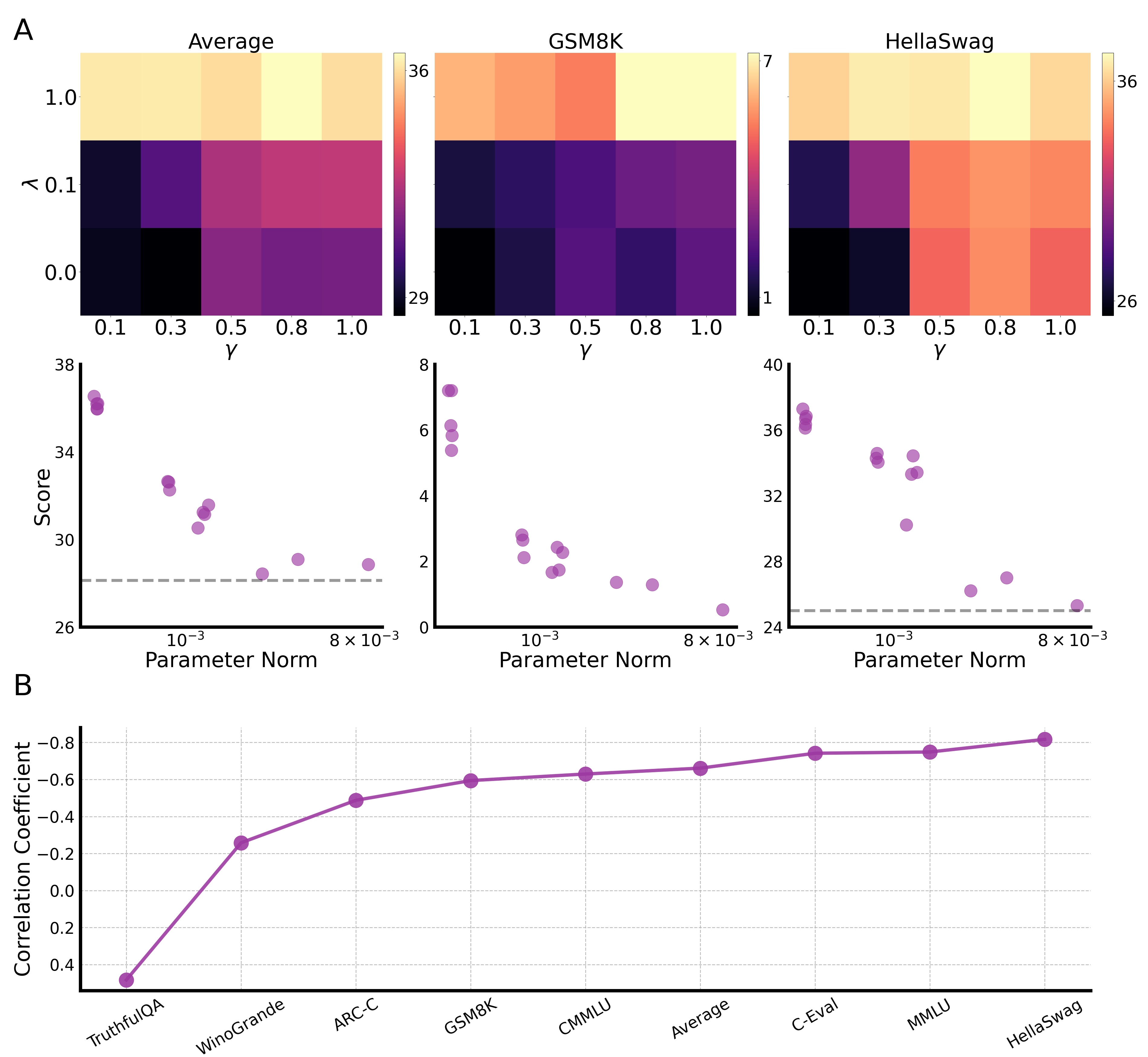}
    \caption{(A) Evaluation scores (average, GSM8K, HellaSwag) under varying complexities. Top: Performance landscape across $\gamma-\lambda$ with color indicating score (dark: low, light: high). Bottom: Score-complexity relationships with points indicating the models and the dashed lines denoting baseline performance levels.
(B) Task-specific Spearman correlations between model complexity and task score. Stronger negative correlations (approaching -1) indicate greater performance enhancement through complexity control.
    }
    \label{fig:187-eval}
\end{figure}

We perform evaluations of models across multiple benchmark tasks. Figure~\ref{fig:187-eval}A depicts the performance distribution of the average score, GSM8K, and HellaSwag, showing better performance as stronger complexity control. 
As shown in Figure \ref{fig:187-eval}B, 
model performance demonstrates a strong inverse correlation with model complexity. 
Full evaluation results are provided in Appendix~\ref{app:187}.

\subsection{Model Analysis}

\begin{figure}[htpb]
    \centering
    \includegraphics[width=1\linewidth]{ 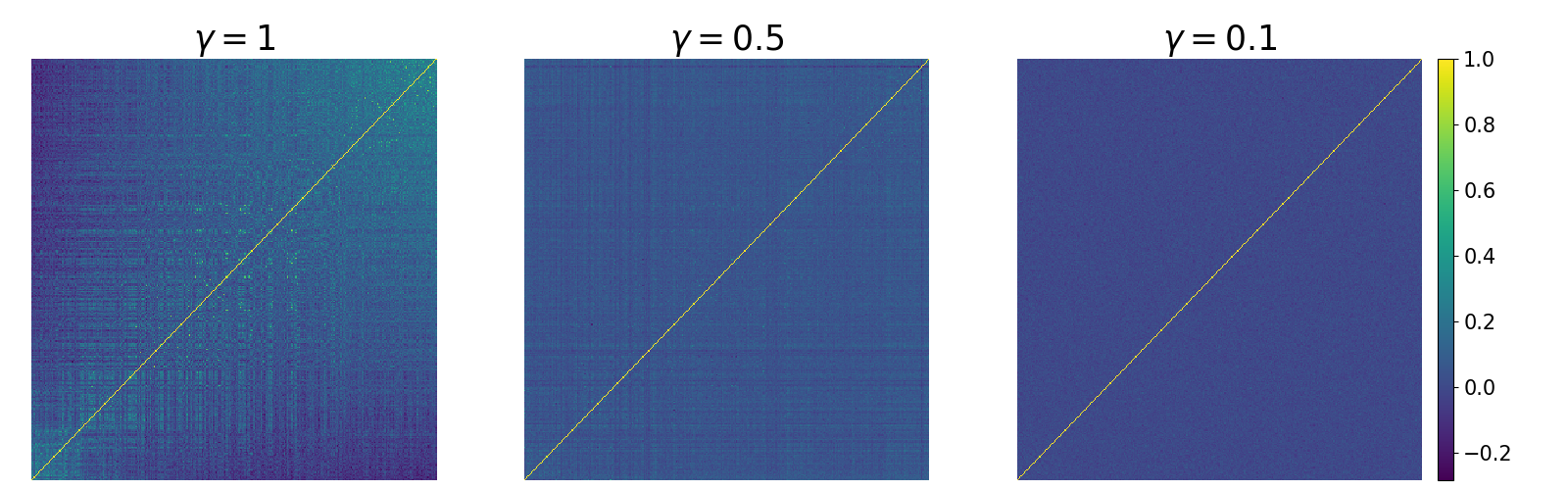}
    \caption{Cosine similarity among 350 embedding vectors which occur most frequently in training dataset under initialization scales $\gamma=1,0.5,0.1$ ($\lambda=0$).}
    \label{fig:187-embedding}
\end{figure}

\paragraph{Embedding space} 
The embedding space reflects the model's representation of the vocabulary and its learning patterns. We compare the cosine similarity of the 350 most frequent embedding vectors under different initialization scales. The results in Figure~\ref{fig:187-embedding} demonstrate that with large complexity ($\gamma=0.1$), the embeddings are pairwise orthogonal, indicating the model ignores their relationship. In contrast, controlling complexity ($\gamma=0.5, 1$) significantly increases the similarity among embeddings, consistent with previous study on condensation phenomenon \cite{luo2021phase}. This analysis suggests that small complexity encourages the model to find associations among tokens.

\begin{figure}[htpb]
    \centering
    \includegraphics[width=1\linewidth]{ 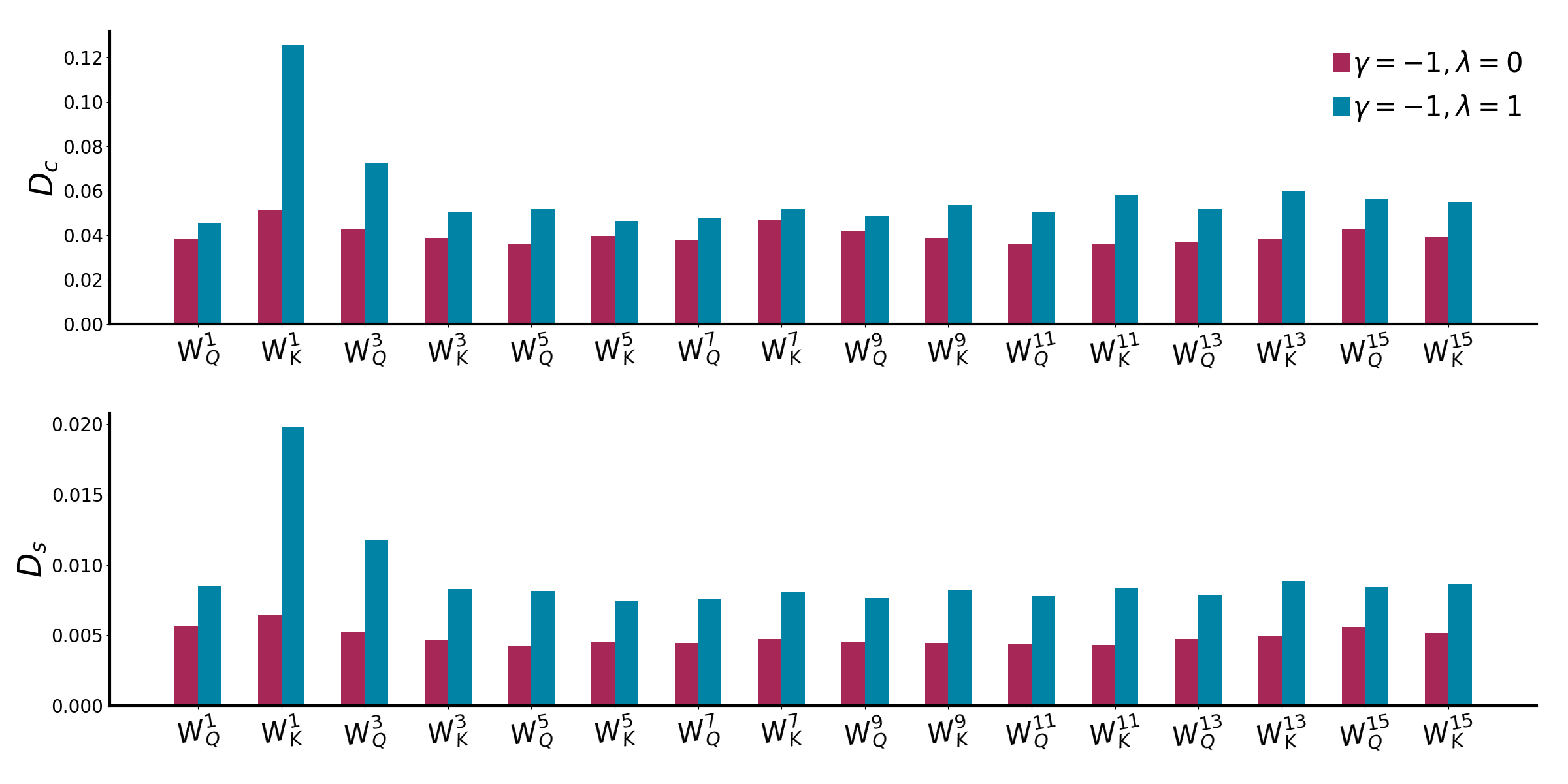}
    \caption{$D_c$ and $D_s$ of $\vW_Q$ and $\vW_K$ in each layer under different model complexity configurations.}
    \label{fig:187-dc}
\end{figure}

\paragraph{Attention matrix}
For each trainable parameter matrix $\vW\in\sR^{d_{in}\times d_{out}}$, we define the following metrics to measure its condensation degree and low-rank degree, respectively.
\begin{equation}
    D_{c}(\vW):= \frac{1}{d_{in}d_{out}}\sum_{i,j}\frac{\vW_i^T\vW_j}{||\vW_i||_2\cdot||\vW_j||_2},\qquad D_s=\frac{\max_i S_{\vW,i}}{\sum_i S_{\vW,i}},
\end{equation}
where $\vW_i$ and $S_{\vW,i}$ means the $i$-th row of $\vW$ and the $i$-th singular value of $\vW$. Larger values of $D_c$ and $D_s$ indicate fewer effective directions in matrix $\vW$, suggesting $\vW$ learns a smaller set of features for fitting dataset. Figure~\ref{fig:187-dc} exhibits the $D_c$ and $D_s$ of the query projection and key projection matrices of models with distinct complexities. The results demonstrate that controlling model complexity effectively increases $D_c$ and $D_s$ across the attention matrices, suggesting that the attention modules in small-complexity models focus more on the fundamental relationships between tokens within sequences, which results in stronger generalization and reasoning ability.


\section{Discussion}

\subsection{Training Stability}
A potential concern in small complexity is training stability, particularly the emergence of loss spikes during optimization. This instability tends to escalate with increasing model scale, as observed in our experiments. To address this challenge in training our 2.4B parameter models, we adopt the $\gamma=0.58$ which is not too large, and implement embedding normalization and sandwich normalization, which successfully mitigate the loss spike phenomenon. A detailed discussion is provided in Appendix~\ref{app:0.58-1}.

\subsection{Training dynamics}
\begin{figure}[htpb]
    \centering
    \includegraphics[width=0.9\linewidth]{ 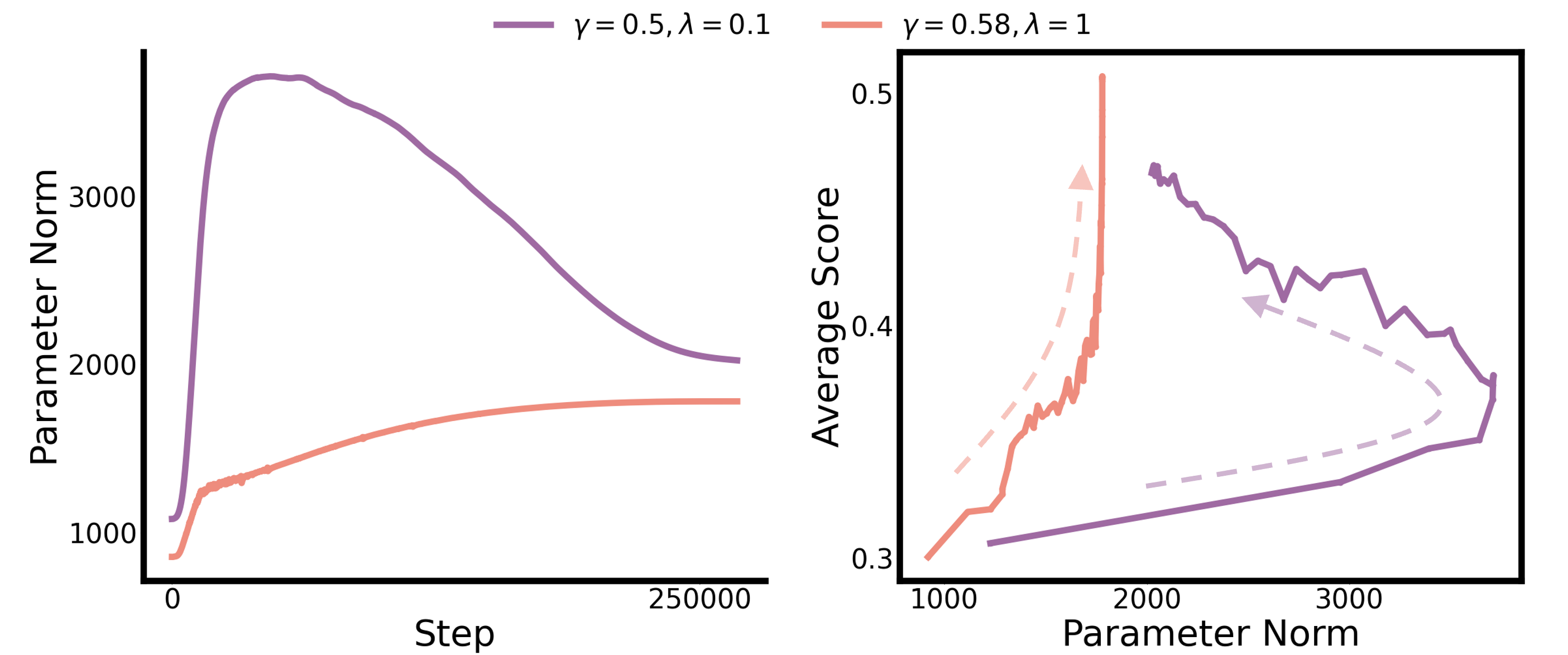}
    \caption{The evolution of parameter norm and average score with two distinct $\gamma-\lambda$ configurations.}
    \label{fig:para-score-evoluation}
\end{figure}

Distinct $\gamma-\lambda$ configurations induce distinct learning dynamics and complexity trajectories, accompanied by different performance evolution patterns. We evaluate the two 2.4B models referenced in Table~\ref{tab:model_benchmark} every 5000 steps during pre-train and obtain the dynamics of the average score. Figure~\ref{fig:para-score-evoluation} characterizes the temporal evolution of model complexity and the average score. For the large-complexity model ($\gamma=0.5, \lambda=0.1$), complexity initially increases followed by progressive decay, during which performance exhibits slow improvement in the ascending phase but accelerates substantially after the complexity turning. Conversely, small-complexity configurations maintain monotonic complexity growth with steady performance gains throughout training. 

\subsection{Theoretical analysis}

This section employs heuristic calculations to analyze how small initialization enhances the generalization of LLMs.

Recall that a 2-layer net $f:\R^d\to\R$ with 1-homogeneous activation (such as ReLU) can be written compactly as $f_{\pi}(\x):=\int \sigma(\w\cdot\x) d\pi(\w)$ for some finite signed measure $\pi$ over the parameter space $\R^d$ (without loss of generality, we can replace $\x$ by $[\x,1]$ to account for the bias) \cite{e_machine_2019,e2022representation}.
There are two common functional spaces for these functions.
In the kernel regime \cite[Theorem 3.9]{e2020comparative}, the function space is characterized by the RKHS norm \cite{rahimi2008uniform}
\begin{equation*}
\|f\|_{\mathcal{H}} = \inf_{\pi} \Big\| \frac{\delta\pi}{\delta\pi_0} \Big\|_{L^2(\pi_0)} = \Big(\int \Big|\frac{\delta\pi}{\delta\pi_0}(\w)\Big|^2 d\pi_0(\w)\Big)^{1/2}, \quad \text{s.t.}~f=f_{\pi} ~\text{and}~ \pi \ll \pi_0
\end{equation*}
where $\pi_0$ is some base distribution (typically the initialization distribution of $\w$) and $\delta\pi/\delta\pi_0$ is the Radon-Nikodym derivative.
In the mean-field regime \cite{mei_mean_2018,rotskoff_parameters_2018}, the function space is characterized by the Barron norm \cite{e2022barron,e2022representation}
\begin{equation*}
\|f\|_{\mathcal{B}} = \inf_{\pi} \|\pi\|_{\text{TV}} = \inf_{\pi} \int d|\pi|(\w), \quad \text{s.t.}~f=f_{\pi}~\text{and}~\text{supp}(\pi)\subseteq\mathbb{S}^d
\end{equation*}
where $\|\cdot\|_{\text{TV}}$ denotes total variation and $\pi$ is supported on the unit sphere $\mathbb{S}^d$.
Naturally, we can define an interpolation between these norms: for any $\gamma > -3/2$,
\begin{align*}
\|f\|_{\gamma}&:= \inf_{\pi} \Big\| \frac{\delta\pi}{\delta\pi_0} \Big\|_{L^{3+2\gamma}(\pi_0)}, \quad \text{s.t.}~f=f_{\pi}~\text{and}~\text{supp}(\pi)\subseteq\mathbb{S}^d
\end{align*}
and by 1-homogeneity $\pi_0$ can be assumed to be supported on $\mathbb{S}^d$ as well.
One can check that $\|\cdot\|_{\gamma=-1/2}=\|\cdot\|_{\mathcal{H}}$ and $\|\cdot\|_{\gamma=-1}=\|\cdot\|_{\mathcal{B}}$.
By Jensen's inequality, the function space of $\|\cdot\|_{\gamma}$ becomes larger as $\gamma$ decreases.
In particular, if $\gamma \leq -1$, then $\pi$ does not have to be absolutely continuous with respect to $\pi_0$, i.e. it can develop singular parts, sometimes known as ``condensation" \cite{CSIAMchen}.
Else, $\gamma>-1$ and $\pi$ must have the form $a\pi_0$.

The question is how $\gamma$ affects deep networks.
Let $\mathcal{M}(\R^d,\R^k)$ denote $\R^k$-vector-valued finite measures over $\R^d$.
For simplicity, consider deep residual nets, $\x^l=\x^{l-1}+f_l(\x^{l-1})$ for $l=1, \dots L$, where each block $f_l:\R^d\to\R^d$ is a 2-layer net parametrized by some $\pi_l \in \mathcal{M}(\R^d,\R^d)$.
For any fixed sequence of input weights $(\w_l)_{l=1}^L \in \R^{L\times d}$, fixed $(\pi_l)$, and an input $\x^0$ drawn from some data distribution, the activation sequence $\big(\sigma(\w_l\cdot\x_l)\big)$ is a stochastic process with a potentially complicated dependency structure.
This is intended to capture cross-layer collaborations in LLMs or ``circuits", such as the induction head \cite{olsson2022incontext}, IOI circuit \cite{wang2023interpretability}, and arithmetic circuits \cite{lindsey2025biology}.
Therefore, we model the parameter distribution of the full network by $\pi\in \mathcal{M}(\R^{Ld},\R^{Ld})$, instead of just the sequence $(\pi_l)$, in order to model the dependency among the activated parameters.
A generalization of the norm $\|\cdot\|_{\gamma}$ from 2-layer nets to deep residual nets can be defined by
\begin{equation*}
\|f\|_{\gamma} = \inf_{\pi} \Big\| \frac{\delta\pi}{\delta\pi_0^{\otimes L}} \Big\|_{L^{3+2\gamma}(\pi_0^{\otimes L})}
= \inf_{\pi} \Big(\int \Big\|\frac{\delta\pi}{\delta\pi_0^{\otimes L}}(\oplus_{l=1}^L\mathbf{w}_l)\Big\|^{3+2\gamma} \prod_{l=1}^L d\pi_0(\mathbf{w}_l) \Big)^{1/(3+2\gamma)}
\end{equation*}
which ranges among all parametrizations $(\pi_l)$ of $f$ and the resulting dependency $\pi$ with respect to the data distribution.
To study $\|\cdot\|_{\gamma}$, we make the following assumptions:
\begin{enumerate}
    \item Among the approximate global minimizers of the loss, training initialized with rate $\gamma$ for any $\gamma>-3/2$ always converges to a minimizer $f^*$ with the minimum $\|\cdot\|_{\gamma}$ norm.
    This assumption holds for 2-layer nets in the kernel regime ($\gamma=-1/2$) when trained by gradient descent \cite{zhang2021understanding}, and we expect similar behavior in general settings.
    
    \item The parameter distribution $\pi^*$ of each approximate global minimizer $f^*$ can always be decomposed into a weighted sum of product measures
    $$\pi^*=\sum_i c_i \pi^i, \quad c_i \in \R, \quad \pi^i=\bigotimes_{l=1}^L \pi^i_l, \quad \|\pi_l^i\|_{\text{TV}}=1$$
    and these $\pi^i$ have disjoint supports (up to $\pi$-negligible subsets).
    Furthermore, the variation of each $\pi^i$, namely $|\pi^i_l|$, is simply $\pi_0$ restricted to some subset $S^i_l\subseteq\R^d$, i.e. $|\pi^i_l| = \pi_0 \mathbf{1}_{S^i_l} / \pi_0(S^i_l)$, and there exists some constant $0<\epsilon\ll 1$ such that either $\pi_0(S^i_l)=\epsilon$ or $\pi_0(S^i_l) = 0.9$.
    The purpose of this assumption is to simplify computation, and we expect our results to hold in much more general settings.
\end{enumerate}
For each $\pi^i$, denote by $L_i$ the number of $\pi^i_l$ with $\pi_0(S^i_l)=\epsilon$.
This $L_i$ can be interpreted as the circuit depth, the number of layers where $\pi^i$ is non-trivial.
The norm of each minimizer $f^*$ becomes
\begin{align*}
\|f^*\|_{\gamma} &= \Big(\sum_i \int \Big\|\frac{\delta(c_i\pi^i)}{\delta\pi_0^{\otimes L}}\Big\|^{3+2\gamma} d\pi_0^{\otimes L} \Big)^{1/(3+2\gamma)} = \Big(\sum_i c_i^{3+2\gamma} \prod_{l=1}^L \int \Big\|\frac{\delta\pi^i_l}{\delta\pi_0}\Big\|^{3+2\gamma} d\pi_0 \Big)^{1/(3+2\gamma)}\\
&= \Big(\sum_i \big( c_i \epsilon^{-L_i} 0.9^{-(L-L_i)} \big)^{3+2\gamma} \Big)^{1/(3+2\gamma)} = 0.9^{-L} \big\| c_i (0.9/\epsilon)^{L_i} \big\|_{l^{3+2\gamma}}
\end{align*}
Hence, each minimizer can be viewed as a circuit ensemble, characterized by its circuit weights and circuit depths $\{(c_i, L_i)\}$.
Recall that for general $l^p$ norms, those with large $p$ are sensitive to the maximum value, while those with small $p$ are sensitive to the amount of non-zero elements.
Thus, for large $\gamma$ such as $-1/2$, the circuit ensemble chosen by training tends to be dense but uniformly shallow (many $c_i>0$ but small $\max L_i$), whereas for small $\gamma$ such as $-1$, the chosen circuits are more likely to be sparse and deep (few $c_i>0$ but $L_i$ can be large).

Intuitively, sparse and deep circuits signify a generalizable solution, as the model has managed to compress the diverse training data into very few but flexible patterns.
Meanwhile, dense and shallow circuits may imply that the model does not have a deep understanding of the data and has to memorize a lot.
In conclusion, it is plausible that small initialization increases generalizability by shifting the preference for circuit ensembles.
Although our calculation is based on simple residual networks, it is reasonable to expect that a similar mechanism applies to Transformers.

\begin{ack}

This work is supported by the National Key R\&D Program of China Grant No. 2022YFA1008200, the National Natural Science Foundation of China Grant No. 92270001, 12371511, 12422119.


\end{ack}


\bibliographystyle{plain}
\bibliography{ref}





\newpage
\appendix


\section{Limitation and Future Work}\label{app:limitation}
While our methodology received thorough empirical validation, scaling to larger models and datasets remains constrained due to computational resource limitations. Future work will prioritize extending the framework to a larger training scale while exploring complexity control mechanisms during post-training.

\section{Experiments Setup}\label{app:setup}


\paragraph{Model architecture} 
Our models are based on the Llama-architecture~\cite{touvron2023llama}. Architectural configurations for different model scales are presented in Table~\ref{tab:compact_config}. For the 2.4B variant, we incorporate embedding normalization and sandwich normalization techniques to enhance training stability.
\begin{table}[htbp]
\centering
\caption{Training Configuration Across Model Scales}
\label{tab:compact_config}
\begin{tabular}{@{}lcccc@{}}
\toprule
\textbf{No. } & \textbf{Model Scale}  & \textbf{Architecture Configuration}  &\textbf{Pre-train Datasize}\\
\midrule
1 & 180M  & 
\begin{tabular}[c]{@{}c@{}}Layers: 16\\ Head Dim: 80\\ Heads (KV): 16 (16)\end{tabular} & 40B  \\

\midrule
2 & 0.9B & 
\begin{tabular}[c]{@{}c@{}}Layers: 32\\ Head Dim: 64\\ Heads (KV): 32 (32)\end{tabular} & 600B\\

\midrule
3 & 2.4B  & 
\begin{tabular}[c]{@{}c@{}}Layers: 44\\ Head Dim: 80\\ Heads (KV): 40 (8)\end{tabular} &1T \\

\midrule

\bottomrule
\end{tabular}
\end{table}

These configurations are systematically mapped as follows: 
\begin{itemize}
    \item \textbf{Configuration 1} governs results in:
    \begin{itemize}
        \item Figure~\ref{fig:token_loss} and~\ref{fig:paranorm}
        \item Section~\ref{sec:analysis-187}
        \item Appendix~\ref{app:187}
    \end{itemize}
    
    \item \textbf{Configuration 2} applies to:
    \begin{itemize}
        \item Figure~\ref{fig:base_sft},~\ref{fig:Dc_0.94} and~\ref{fig:Ds_0.94}
        \item Columns ``0.9B Large'' and ``0.9B Small'' in Table~\ref{tab:model_benchmark},~\ref{tab:base_benchmark}
    \end{itemize}
    
    \item \textbf{Configuration 3} encompasses:
    \begin{itemize}
        \item Columns ``2.4B Large'' and ``2.4B Small'' in Table~\ref{tab:model_benchmark} and~\ref{tab:base_benchmark}
        \item Figure~\ref{fig:para-score-evoluation},~\ref{fig:Dc_2.4} and~\ref{fig:Ds_2.4}
        \item Appendix~\ref{app:0.58-1}
    \end{itemize}
\end{itemize}
For Figure~\ref{fig:scalinglaw}, the corresponding model specifications are detailed in Table~\ref{tab:scaling_config}.
\begin{table}[htbp]
\centering
\caption{Architecture Configuration adopted in Figure~\ref{fig:scalinglaw}.}
\label{tab:scaling_config}
\begin{tabular}{@{}lcccc@{}}
\toprule
 \textbf{Model Scale (M)}  & \textbf{Layers}  &\textbf{Heads (KV)} &\textbf{Head Dim}\\
\midrule
50  & 12 & 12 & 64 \\

\midrule
 100 &  16 & 16 & 60\\

\midrule
200 & 24 & 18 & 60 \\

\midrule
400 & 24 & 24 & 64 \\

\midrule
800 & 32 & 24 & 80 \\

\bottomrule
\end{tabular}
\end{table}

\paragraph{DataSet}
The 180M models are trained on a carefully curated 40B subset obtained through uniform sampling from the Memory$^3$ training corpus. The 0.9B models undergo pre-training on a 600B subset derived from the SlimPajama corpus. Building upon the Memory$^3$ foundation, we expand the training corpus through a rigorous data processing pipeline that integrates deduplication protocols, multi-dimensional quality assessment metrics, and optimized domain ratio adjustments. This systematic curation process yields a refined 1T training dataset, which is the basis for training our 2.4B models. For the SFT data, we also adopt the SFT data from Memory$^3$.

\paragraph{Training setup}
The training is implemented using Microsoft's Megatron-DeepSpeed framework~\cite{Megatron-DeepSpeed}, utilizing a mixed-precision configuration where model parameters, gradients, and activations were maintained in bfloat16 format while preserving optimizer states in float32 precision for the AdamW optimizer. The learning rate schedule adopted a cosine annealing strategy with linear warmup, where the warmup phase spanned the initial 5\% of the total training iterations. The learning rate boundaries were configured with maximum and minimum values of $1\times10^{-3}$ and $1 \times 10^{-5}$.

\paragraph{Training cost}
We present the computational costs for single training sessions across different model scales. The 180M model required 32 MX-C500 accelerators with a training duration of 12 hours. For the 0.9B architecture, the training process utilized 400 MX-C500 accelerators over 72 hours. 512 MX-C500 accelerators are employed for training a 2.4B model in one week.

\paragraph{Evaluation Details}
Our model are assessed across multiple open-source benchmarks via lm-eval-harness~\cite{eval-harness}, covering areas such as factual knowledge: CMMLU~\cite{li2023cmmlu}, C-Eval~\cite{huang2023ceval}, MMLU~\cite{mmlu} and its enhanced version MMLU-Pro~\cite{mmlu_pro}, OpenBookQA~\cite{OpenBookQA2018}, and GPQA~\cite{rein2024gpqa}. Language comprehension: BBH~\cite{bbh}, ARC-C~\cite{arc}, TruthfulQA~\cite{lin2021truthfulqa}, WinoGrande~\cite{ai2:winogrande}, HellaSwag~\cite{zellers2019hellaswag}, AGIEval-EN~\cite{zhong2023agieval}, CommonsenseQA~\cite{talmor-etal-2019-commonsenseqa}. Code generation: MBPP~\cite{mbpp} and IFEval~\cite{ifeval}. Mathematical reasoning: MATH~\cite{MATH} and GSM8K~\cite{cobbe2021gsm8k}.

\section{Evaluation of base models}\label{app:base}
Table~\ref{tab:base_benchmark} presents the evaluation results of the base models. The results reveal that the 0.9B model with small complexity does not demonstrate a significant advantage, while the 2.4B small-complexity model achieves measurable improvements. This disparity potentially stems from differences in pre-training data quality. Specifically, the 0.9B model's training data (subsampled from the SlimPajama corpus) exhibits suboptimal quality, limiting its downstream task adaptability.
\begin{table}[htbp]
\centering
\vspace{-5pt}
\begin{threeparttable}
\caption{Evaluation of base models}
\label{tab:base_benchmark}
\begin{tabular}{@{}lccccc@{}}
\toprule
\textbf{Models\tnote{1}} & \textbf{0.9B Large} & \textbf{0.9B Small} & \textbf{2.4B Large} & \textbf{2.4B Small}\\
$\left(\gamma,\lambda\right)$ & (0.5,0.1) &(1,1) &(0.5,0.1) & (0.58,1)\\
\midrule
MMLU          & 25.7  & 24.4 (\mcolor{-1.3})  & 36.5  & 41.1 (\ppcolor{+4.6}) \\
MMLU-Pro      & 11.8  & 11.3 (\mcolor{-0.5})  & 13.4  & 14.9 (\pcolor{+1.5}) \\
BBH           & 31.1  & 31.3 (\pcolor{+0.2})  & 34.7  & 35.7 (\pcolor{+1.0}) \\
ARC-C         & 34.6  & 35.6 (\pcolor{+1})  & 39.9     & 47.0 (\ppcolor{+7.1}) \\
TruthfulQA    & 36.6  & 36.8 (\pcolor{+0.2})  & 36.6     & 41.1 (\ppcolor{+4.5}) \\
WinoGrande   & 59.4  & 61.1 (\pcolor{+1.7}) & 64.0 & 68.0 (\ppcolor{+4.0}) \\
HellaSwag     & 59.4  & 62.0 (\pcolor{+2.6}) & 66.5     & 67.8 (\pcolor{+1.3}) \\
AGIEval-EN    & 17.8 & 17.8  & 20.5 & 19.9 (\mcolor{-0.6})\\

OpenBookQA   &39   &37.8 (\mcolor{-1.2}) & 42.2 & 41.6 (\mcolor{-0.6})\\

CommonsenseQA  &19   &20.9 (\pcolor{+1.9}) & 38.2  & 48.7 (\ppcolor{+10.5}) \\

GPQA          & 23.7  & 24.6 (\pcolor{+0.9})  & 24.6     & 24.8 (\pcolor{+0.2}) \\
MATH          & 2.0  & 2.5 (\pcolor{+0.5})  & 30.0  & 29.4 (\mcolor{-0.6}) \\
GSM8K         & 1.9  & 2.4 (\pcolor{+0.5}) & 31.0  & 39.7 (\ppcolor{+8.7}) \\

MBPP          & 0.2  & 6.2 (\ppcolor{+6})  & 19.2  & 22 (\pcolor{+2.8})  \\
IFEval        &27.2  & 27.7 (\pcolor{+0.5}) & 27.0 & 28.4 (\pcolor{+1.4})\\
\bottomrule
\end{tabular}
\begin{tablenotes}
    \footnotesize
        \item[1] ``Large'' and ``Small'' mean large complexity and small complexity, respectively. 
\end{tablenotes}
\end{threeparttable}
\end{table}
\newpage

\section{Training Stability}\label{app:0.58-1}
\begin{figure}[htpb]
    \centering
    \includegraphics[width=1\linewidth]{ 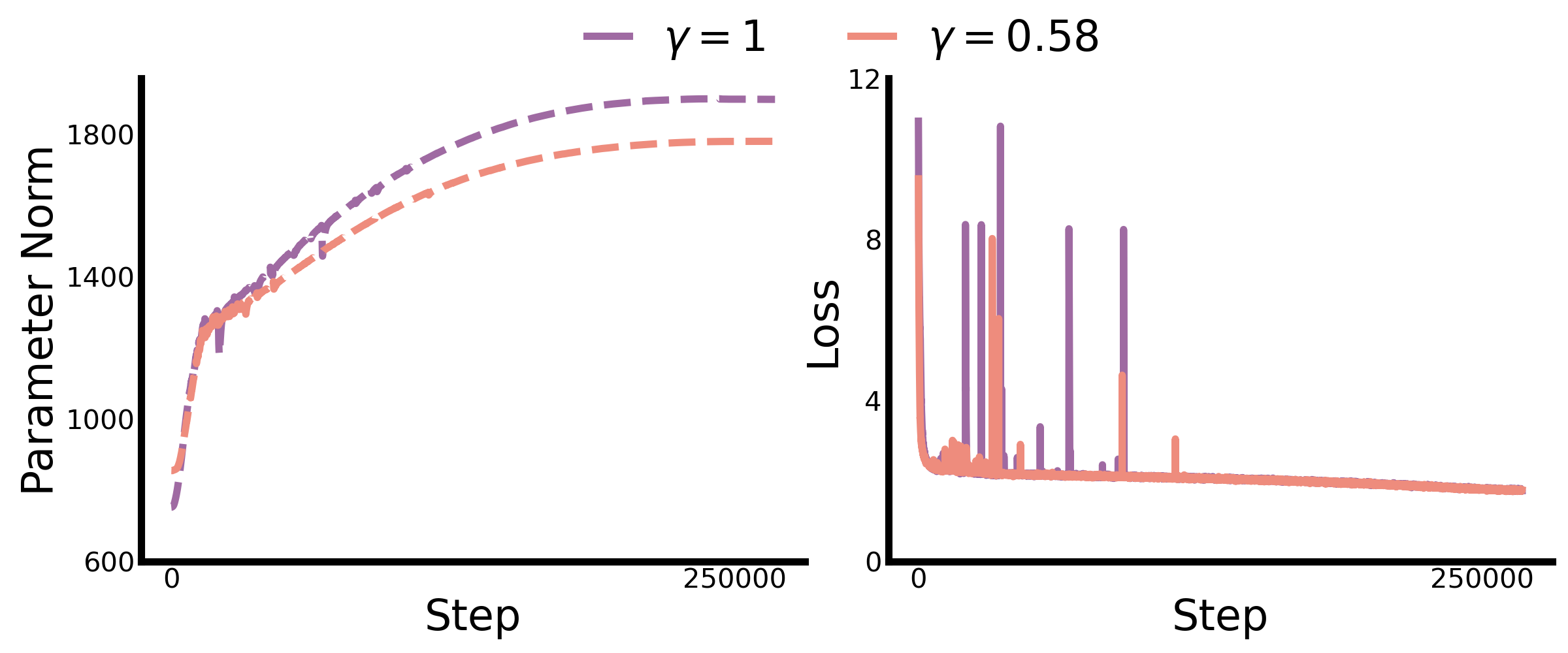}
    \caption{The dynamics of parameter norm (left) and loss (right) of 2.4B model with $\gamma=1,\lambda=1$ (purple) and $\gamma=0.58,\lambda=1$ (orange).}
    \label{fig:58-1}
\end{figure}

We observe that controlling model complexity induces training instability with loss spikes as the model scale increases, particularly when employing extremely small initialization scales. The purple lines in Figure~\ref{fig:58-1} demonstrate the training dynamics (loss and parameter norm evolution) of the 2.4B model with $\gamma=1, \lambda=1$, revealing severe instability that prevents convergence to small-complexity solutions. Contrastingly, Figures~\ref{fig:paranorm},~\ref{fig:187-eval} establish that sufficiently large $\lambda$ values ($\lambda=1$) enable small-complexity convergence regardless of initialization scale. This insight motivates our strategy: combining large $\lambda$ with critical initialization scaling. Inspired by the initialization practices of GPT-2 and DeepSeek-V3, we adopt $\gamma=0.58$ in our 2.4B model's training. As shown in Figure~\ref{fig:58-1} Table 7, this configuration achieves stabilized training and better performance.

\begin{table}[htbp]
\centering
\caption{Evaluation of 2.4B models with $\gamma=1$ and $\gamma=0.58$.}
\label{tab:0.58-1}
\begin{tabular}{@{}lccc@{}}
\toprule
\makebox[0.1\textwidth]{\textbf{$\gamma$}} & \makebox[0.1\textwidth][c]{1} & \makebox[0.1\textwidth][c]{0.58}\\
\midrule
MMLU          & 65.4    & 64.4  \\
MMLU-Pro      & 30.1   & 30.1  \\
BBH           & 45.0   & 43.7  \\
ARC-C         & 59.0  & 60.7  \\
TruthfulQA    & 64.1   & 61.7 \\
WinoGrande   & 78.0  & 76.9 \\
HellaSwag     & 74.6   & 75.4 \\
AGIEval-EN    & 32.2 & 35.6\\

OpenBookQA    &40.8 & 43.4  \\

CommonsenseQA  &77.7  & 76.9 \\

GPQA          & 29.5   & 31.7 \\
MATH          & 33.3 & 35.3 \\
GSM8K         & 62.0  & 63.8 \\

MBPP          & 21.6  & 22.4  \\
IFEval        & 42.2  & 40.9\\
\midrule
Average       & 50.4  & 50.9\\
\bottomrule
\end{tabular}
\end{table}
\newpage

\section{Evaluation of the 180M models}\label{app:187}
\begin{figure}[htpb]
    \centering
    \includegraphics[width=0.8\linewidth]{ 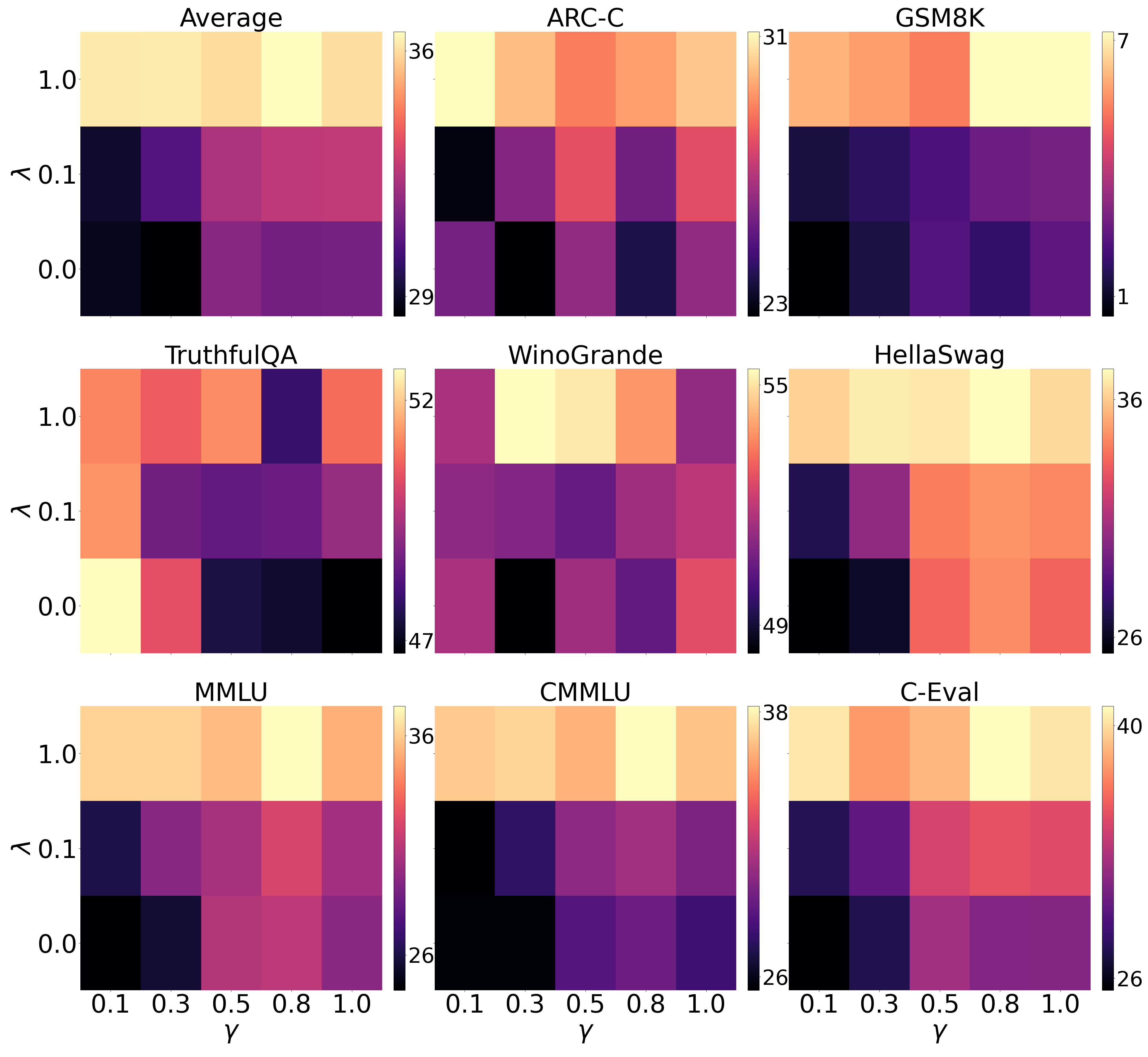}
    \caption{Performance landscape of all tasks across $\gamma-\lambda$ of the 180M models. The color indicates the score (dark: low, light: high).}
    \label{fig:187_eval_appendix1}
\end{figure}

\begin{figure}[htpb]
    \centering
    \includegraphics[width=0.8\linewidth]{ 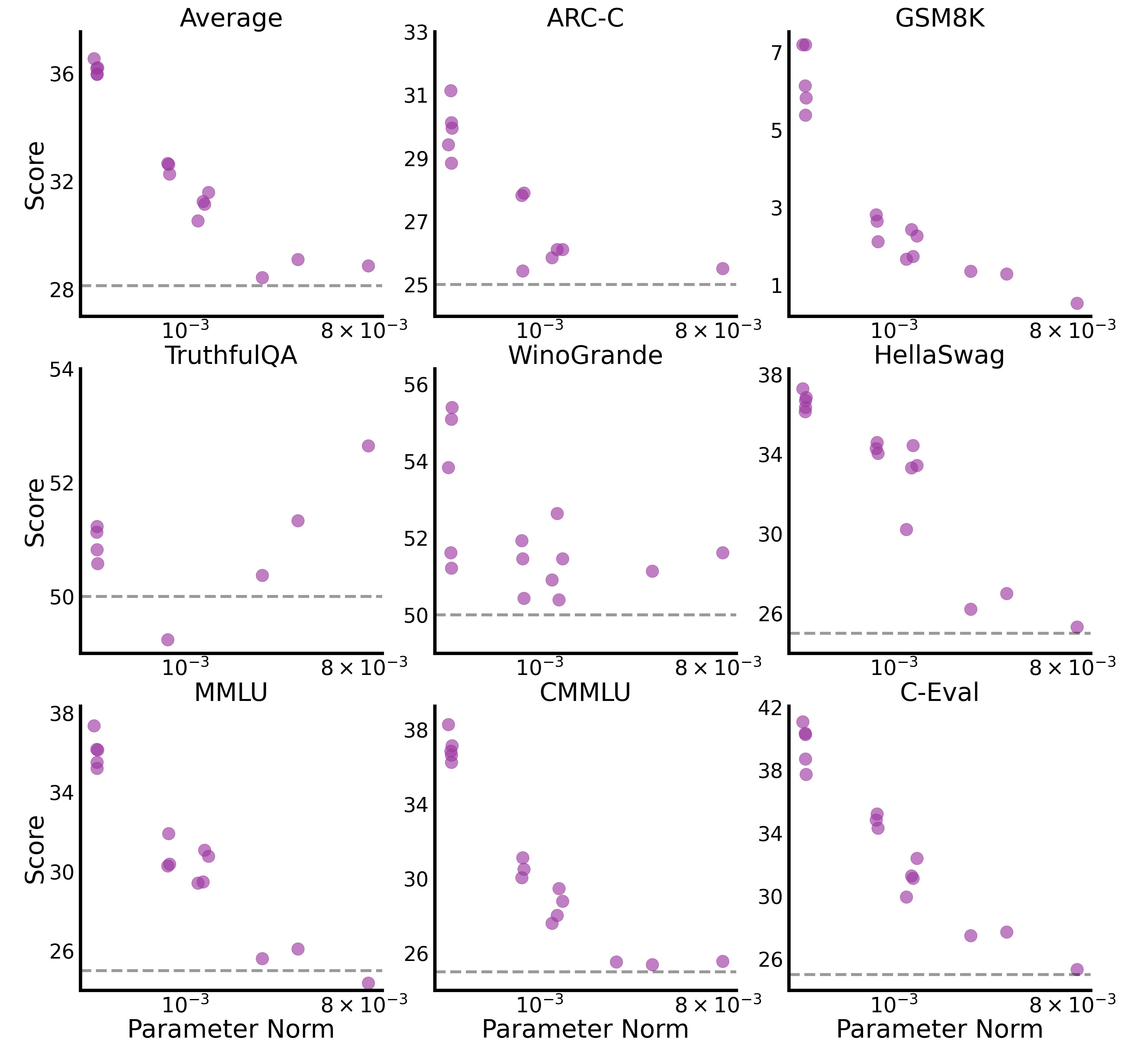}
    \caption{Score-complexity relationship with dashed lines denoting baseline performance levels.}
    \label{fig:187_eval_appendix2}
\end{figure}

Table~\ref{tab:187M_results} comprehensively evaluates all 180M models referenced in Section~\ref{sec:187m}, systematically illustrating performance improvements through complexity control. Additionally, Figure~\ref{fig:187_eval_appendix1}, \ref{fig:187_eval_appendix2} visualizes the performance-complexity correlation across all tasks, further validating our conclusions.

\begin{table}[htbp]
\centering
\caption{Evaluation results of the 180M models under varying $\gamma-\lambda$ configurations}
\label{tab:187M_results}
\footnotesize  
\begin{tabular}{ccccccccccc}
\toprule
\multirow{3}{*}{$\lambda$} & \multirow{3}{*}{$\gamma$} & \multirow{3}{*}{Average} & \multicolumn{8}{c}{Tasks} \\
\cmidrule(lr){4-11}
& & & ARC-C & GSM8K & TruthfulQA & WinoGrande & HellaSwag & MMLU & CMMLU & C-EVAL\\
\midrule
\multirow{5}{*}{0} 
& 0.1 & 30.0 & 25.5 & 0.5 & 52.7 & 51.6 & 25.3 & 24.4 & 25.6 & 25.5 \\
& 0.3 & 29.1 & 22.6 & 1.4 & 50.4 & 48.3 & 26.2 & 25.6 & 25.5 & 27.5 \\
& 0.5 & 31.9 & 26.1 & 2.3 & 47.4 & 51.5 & 33.4 & 30.8 & 28.8 & 32.4 \\
& 0.8 & 31.4 & 23.7 & 1.7 & 47.3 & 50.4 & 34.4 & 31.1 & 29.5 & 31.1 \\
& 1.0 & 31.8 & 26.1 & 2.4 & 46.7 & 52.6 & 33.3 & 29.5 & 28.0 & 31.3 \\
\cmidrule(lr){1-11}

\multirow{5}{*}{0.1}
& 0.1 & 30.0 & 22.9 & 1.3 & 51.3 & 51.1 & 27.0 & 26.1 & 25.4 & 27.7 \\
& 0.3 & 31.1 & 25.9 & 1.7 & 48.7 & 50.9 & 30.2 & 29.4 & 27.6 & 29.9 \\
& 0.5 & 32.2 & 27.9 & 2.1 & 48.5 & 50.4 & 34.0 & 30.4 & 30.5 & 34.3 \\
& 0.8 & 32.4 & 25.4 & 2.7 & 48.6 & 51.5 & 34.6 & 31.9 & 31.1 & 35.2 \\
& 1.0 & 32.7 & 27.8 & 2.8 & 49.2 & 51.9 & 34.3 & 30.3 & 30.1 & 34.8 \\
\cmidrule(lr){1-11}

\multirow{5}{*}{1.0}
& 0.1 & 35.4 & 31.1 & 6.1 & 51.1 & 51.6 & 36.1 & 36.2 & 36.9 & 40.3 \\
& 0.3 & 35.8 & 30.0 & 5.8 & 50.6 & 55.4 & 36.9 & 36.1 & 37.1 & 37.7 \\
& 0.5 & 35.5 & 28.8 & 5.4 & 51.2 & 55.1 & 36.7 & 35.5 & 36.3 & 38.7 \\
& 0.8 & 35.5 & 29.4 & 7.2 & 47.9 & 53.8 & 37.3 & 37.4 & 38.3 & 41.1 \\
& 1.0 & 35.2 & 30.1 & 7.2 & 50.8 & 51.2 & 36.3 & 35.2 & 36.6 & 40.3 \\
\bottomrule
\end{tabular}
\end{table}

\section{Model Analysis}\label{app:parameter}

\subsection{Embedding Structure of 0.9B and 2.4B models}
\begin{figure}[htpb]
    \centering
    \includegraphics[width=1\linewidth]{ 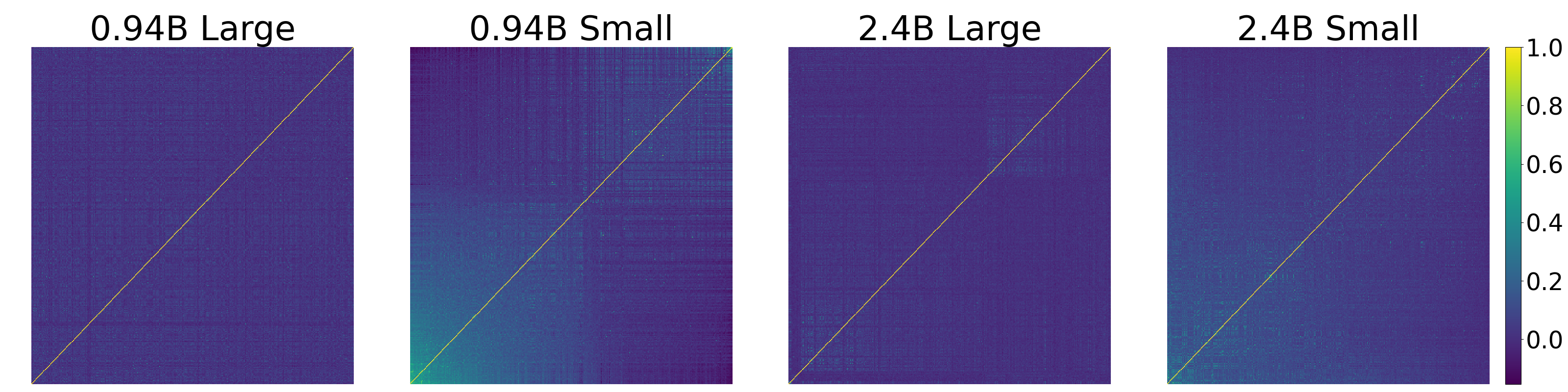}
    \caption{Cosine similarity among 350 embedding vectors which occur most frequently in training dataset under different complexities of 0.9B models and 2.4B models. The "Large" and "small" mean the large complexity and small complexity.}
    \label{fig:embedding_appendix}
\end{figure}

Figure~\ref{fig:embedding_appendix} depicts the cosine similarity among 350 embedding vectors of 0.9B models and 2.4B models, with different model complexities. The results present a similar phenomenon with Figure~\ref{fig:187-embedding}, demonstrating that complexity control contributes to a focus on the association among different tokens.

\subsection{Attention module of 0.9B and 2.4B models}
\begin{figure}[htpb]
    \centering
    \includegraphics[width=0.9\linewidth]{ 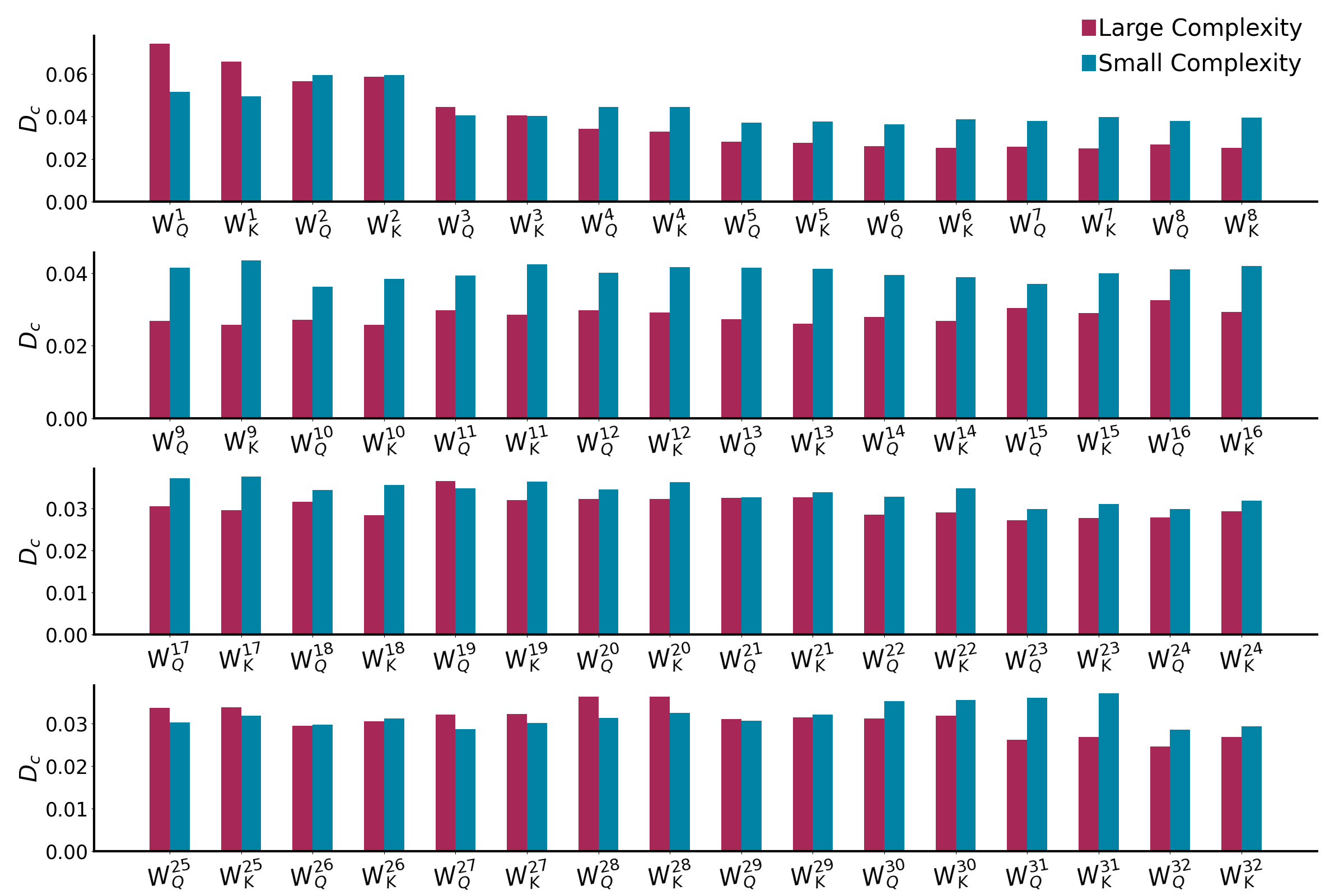}
    \caption{$D_c$ of $\vW_Q$ and $\vW_K$ in 0.9B model's each layer under different model complexity configurations.}
    \label{fig:Dc_0.94}
\end{figure}

\begin{figure}[htpb]
    \centering
    \includegraphics[width=0.9\linewidth]{ 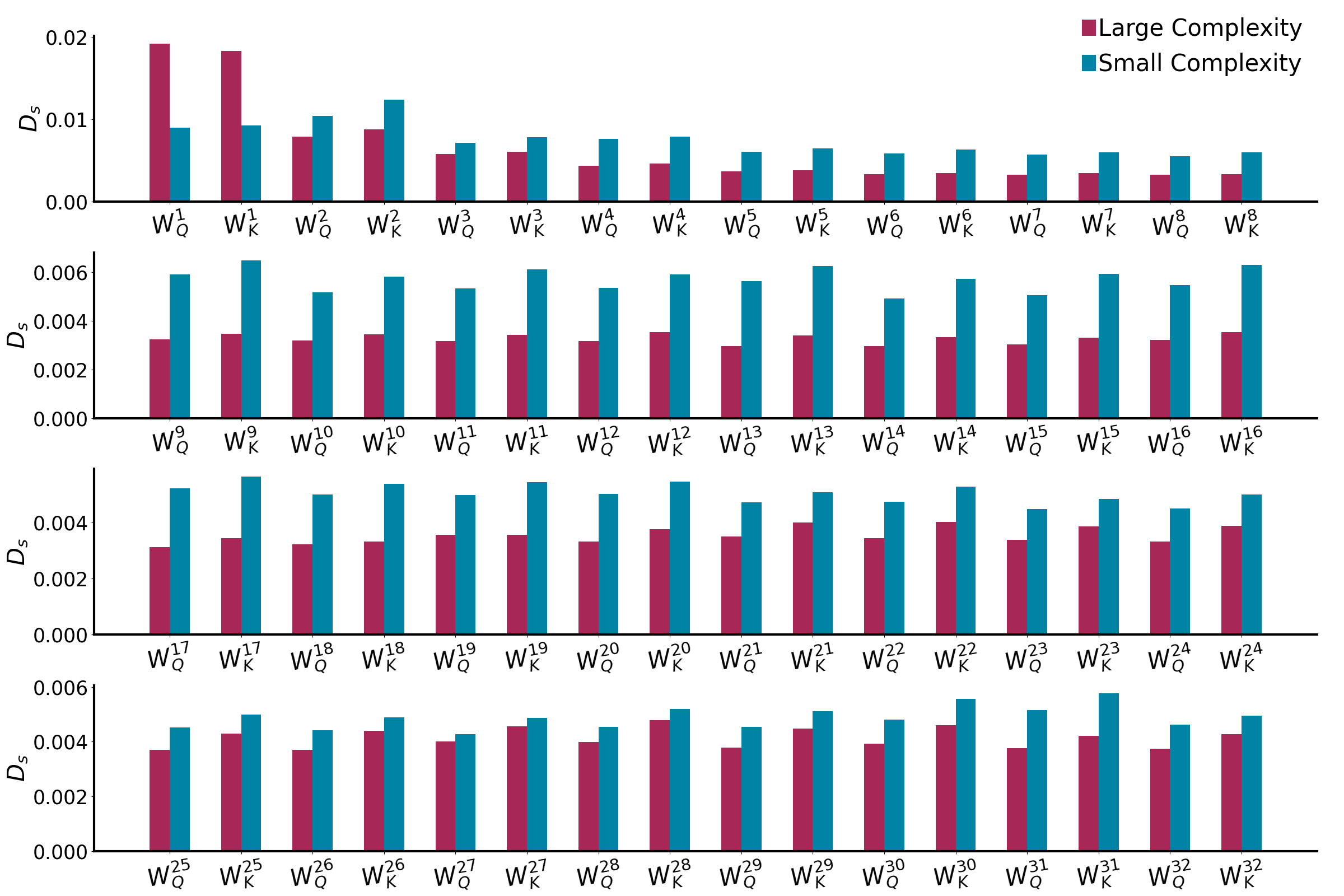}
    \caption{$D_s$ of $\vW_Q$ and $\vW_K$ in 0.9B model's each layer under different model complexity configurations.}
    \label{fig:Ds_0.94}
\end{figure}

\begin{figure}[htpb]
    \centering
    \includegraphics[width=0.9\linewidth]{  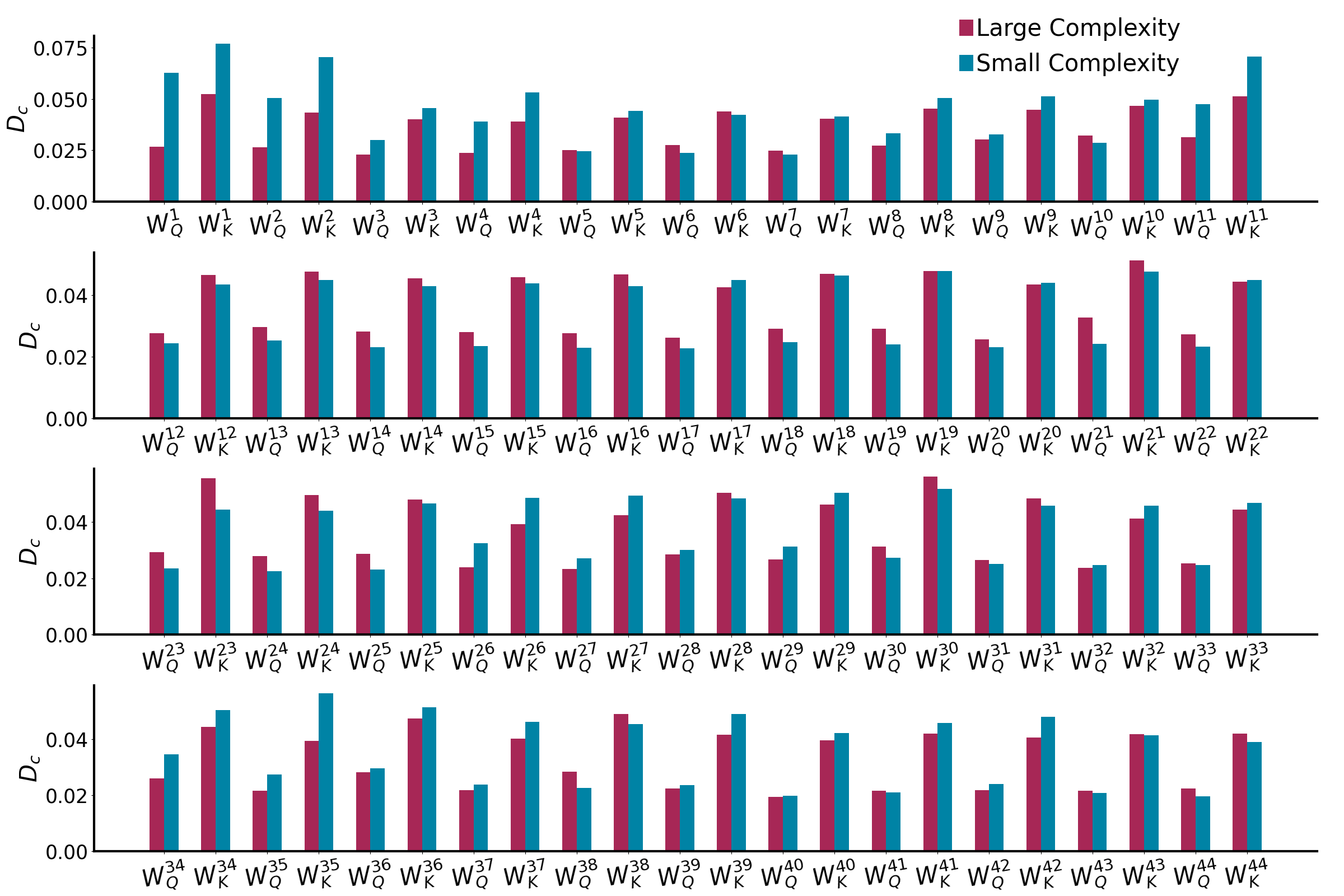}
    \caption{$D_c$ of $\vW_Q$ and $\vW_K$ in 2.4B model's each layer under different model complexity configurations.}
    \label{fig:Dc_2.4}
\end{figure}

\begin{figure}[htpb]
    \centering
    \includegraphics[width=0.9\linewidth]{ 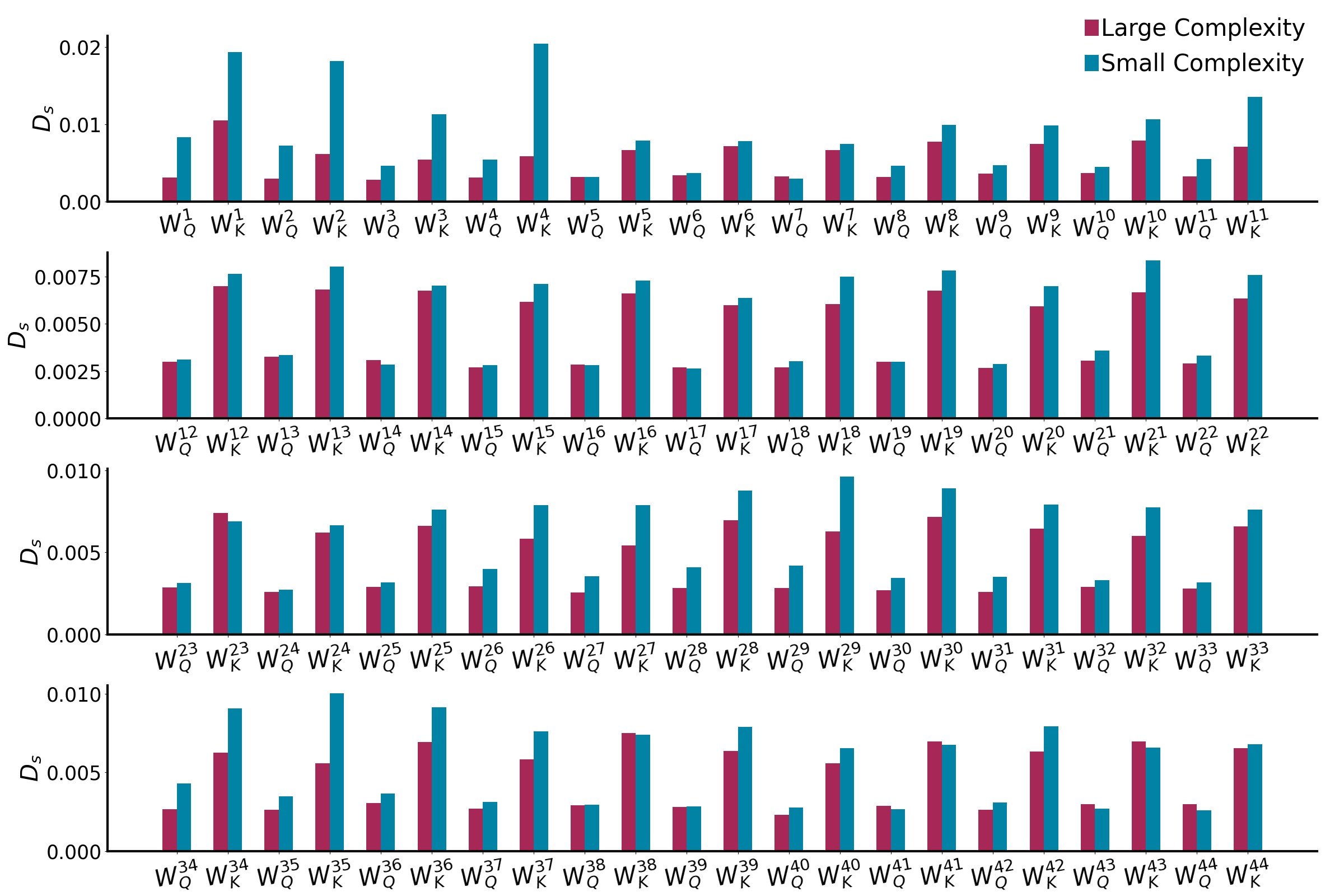}
    \caption{$D_s$ of $\vW_Q$ and $\vW_K$ in 2.4B model's each layer under different model complexity configurations.}
    \label{fig:Ds_2.4}
\end{figure}

Figure~\ref{fig:Dc_0.94},~\ref{fig:Ds_0.94},~\ref{fig:Dc_2.4}, and~\ref{fig:Ds_2.4} exhibit the $D_c$ and $D_s$ of query projection matrices and key projection matrices in the 0.9B models and 2.4B models, reveals a condensation and low-rank trend of the small-complexity models.

\end{document}